\documentclass[review]{elsarticle}
\usepackage{url}
\usepackage{color}
\usepackage{booktabs}
\usepackage{multirow}
\usepackage{multicol}
\usepackage{graphicx}
\usepackage{algorithm}
\usepackage{algorithmic}
\usepackage{amsmath,graphicx}
\usepackage[colorlinks,
            linkcolor=red,
            anchorcolor=blue,
            citecolor=green
            ]{hyperref}
\journal{Journal of Neurocomputing}

\begin{document}

\begin{frontmatter}

\title{Sketch-Specific Data Augmentation for Freehand Sketch Recognition}

\author{Ying Zheng$^{1,2}$}
\author{Hongxun Yao$^{2}$}
\author{Xiaoshuai Sun$^{3}$, Shengping Zhang$^{4}$, Sicheng Zhao$^{5}$, Fatih Porikli$^{6}$}
\address{$^{1}$ Artificial Intelligence Research Institute, Zhejiang Lab, Hangzhou, China\\
$^{2}$ School of Computer Science and Technology, Harbin Institute of Technology, Harbin, China\\
$^{3}$ School of Informatics, Xiamen University, Xiamen, China\\
$^{4}$ School of Computer Science and Technology, Harbin Institute of Technology, Weihai, China\\
$^{5}$ School of Software, Tsinghua University, Beijing, China\\
$^{6}$ Research School of Engineering, Australian National University, ACT, Australia\\
zhengyinghit@outlook.com, \{h.yao, s.zhang\}@hit.edu.cn, xssun@xmu.edu.cn, zhaosicheng@tsinghua.edu.cn, fatih.porikli@anu.edu.au}

\begin{abstract}
Sketch recognition remains a significant challenge due to the limited training data and the substantial intra-class variance of freehand sketches for the same object. Conventional methods for this task often rely on the availability of the temporal order of sketch strokes, additional cues acquired from different modalities and supervised augmentation of sketch datasets with real images, which also limit the applicability and feasibility of these methods in real scenarios.

In this paper, we propose a novel sketch-specific data augmentation (SSDA) method that leverages the quantity and quality of the sketches automatically. From the aspect of quantity, we introduce a Bezier pivot based deformation (BPD) strategy to enrich the training data. Towards quality improvement, we present a mean stroke reconstruction (MSR) approach to generate a set of novel types of sketches with smaller intra-class variances. Both of these solutions are unrestricted from any multi-source data and temporal cues of sketches. Furthermore, we show that some recent deep convolutional neural network models that are trained on generic classes of real images can be better choices than most of the elaborate architectures that are designed explicitly for sketch recognition. As SSDA can be integrated with any convolutional neural networks, it has a distinct advantage over the existing methods. Our extensive experimental evaluations demonstrate that the proposed method achieves the state-of-the-art results (84.27\%) on the TU-Berlin dataset, outperforming the human performance by a remarkable 11.17\% increase. Finally, more experiments show the practical value of our approach for the task of sketch-based image retrieval.
\end{abstract}


\end{frontmatter}

%\linenumbers

\section{Introduction}
\label{sec:introduction}

Sketch recognition has attracted considerable interest over the past decade due to its immediate applications in image retrieval \cite{zhang2016sketch} \cite{wang2015sketchretrieval} and synthesis \cite{chen2009sketch2photo} \cite{sangkloy2016scribbler}, 3D shape retrieval \cite{shao2011discriminative} \cite{wang2015sketch} and reconstruction \cite{Masry2005A} \cite{xu2013sketch2scene}. One of the main differences between sketch recognition and object recognition is that freehand sketch images are lack of prominent color and texture cues, spatially distorted, and highly abstract, which makes sketch recognition a remarkable challenge. Recently, a number of efforts have been devoted to developing effective sketch recognition approaches, which mainly focus on integrating handcrafted features in traditional object recognition frameworks \cite{schneider2014sketch} \cite{li2015free}. Although these methods report certain advancements, their recognition rates still need to be improved for real applications.

\begin{figure}[t]
    \centering
    \centerline{\includegraphics[width=1.0\linewidth]{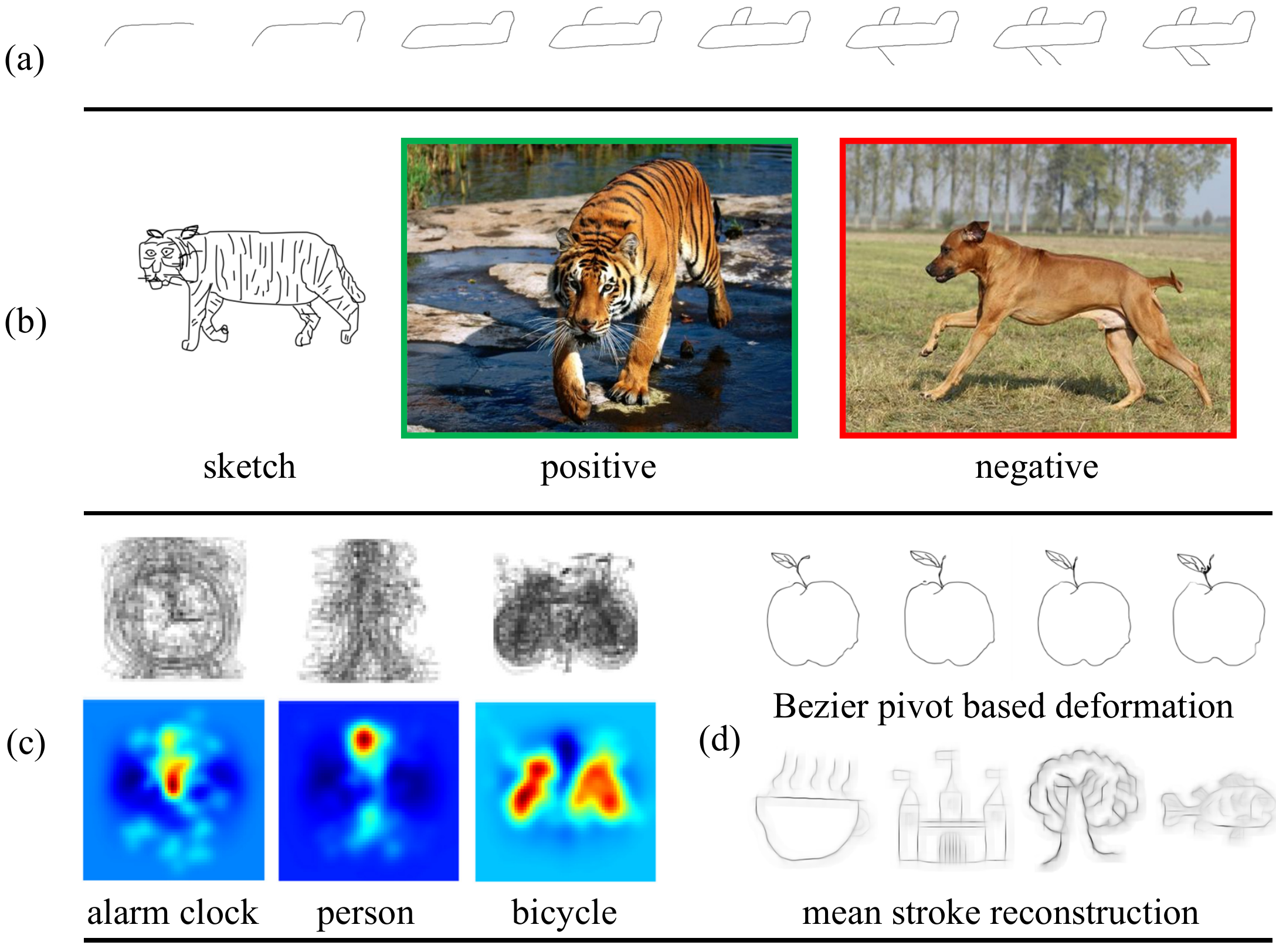}}
    \caption{Illustration of the leading methods for sketch recognition. Some of the existing state-of-the-art methods (a) heavily rely on the temporal order information of human sketching \cite{yu2017sketch}, (b) introduce a large number of real images to construct triplets \cite{zhang2016sketchnet}, or (c) utilize multi-source data like eye fixations \cite{sarvadevabhatla2017object}. These assumptions greatly limit their application range and renders them infeasible for real implementations. In contrast, (d) our approach can achieve a superior performance without any need for temporal information of strokes or multi-source data.}
    \label{fig:motivation}
\end{figure}

In recent years, convolutional neural network (CNN) based methods have revolutionalized the field of object recognition \cite{krizhevsky2012imagenet} \cite{simonyan2014very} \cite{he2016deep}. Intuitively, the CNN models that are pre-trained on real image datasets such as ImageNet \cite{deng2009imagenet} can be transferred directly to the task of sketch recognition. However, this would degrade the recognition performance drastically on sketch datasets due to two reasons: 1) the existing sketch datasets used to fine-tune the pre-trained CNN models are much smaller than their real image counterparts, and 2) the intra-class variance of sketch images is more difficult to model because of the high-level abstraction, which causes discriminative information to be diluted.

A promising solution to the first problem is to apply data augmentation on the sketch datasets, which has been adopted by many researchers. The Sketch-a-Net 2.0 \cite{yu2017sketch}, a representative method for sketch recognition, introduces two sketch-domain specific strategies to augment the training data: sketch removal and sketch deformation. After training on the augmented dataset, its model attains an improved recognition performance. Since this augmentation method relies strongly on the temporal order information of human strokes in sketch generation process as shown in Figure \ref{fig:motivation}(a), its applicability is limited to the devices such as touchpads that can record individual temporal entries. Therefore, this method cannot be used in applications that employ a single static sketch, such as data retrieval by taking a picture of a sketch.

To address the second problem, Zhang et al. \cite{zhang2016sketchnet} propose to learn a shared embedding structure among triplets constructed by a large number of real images as shown in Figure \ref{fig:motivation}(b), which makes their model complicated, thus hard to train. Moreover, feeding such a large number of triplets into a network for sketch recognition is time-consuming. Other approaches attempt to improve the recognition power by incorporating multi-source data, e.g., eye fixations \cite{sarvadevabhatla2017object} (shown in Figure \ref{fig:motivation}(c)), text and clip arts \cite{sun2012query}. They depend on the availability of additional data, which is cumbersome and expensive to collect.

To address the above shortcomings, here we focus on enhancing the quantity and quality of the sketch data by investigating sketch-specific data augmentation (SSDA) solutions. With respect to the quantity, we propose a Bezier pivot based deformation (BPD) approach to generate a substantial amount of new freehand sketches. This BPD approach directly applies to the original single image sketches without requiring temporal cues of sketch lines. Being not subject to the type of input sketch data, BPD enables a broader range of applications. To improve the quality of sketches, we introduce a novel method called mean-stroke reconstruction (MSR) to produce an innovative form of sketches. The MSR uses the mean strokes computed on the training set to reconstruct the original sketches. It can effectively decrease the intra-class variance between freehand sketches. Since it does not demand a large number of same-class real images or rely on any additional cues, it requires low computational complexity when training the CNN model and relieves the cost of data collection.

To provide a more objective and comprehensive evaluation of different CNN based methods, we select 6 widely used CNN models including AlexNet \cite{krizhevsky2012imagenet}, VGG \cite{simonyan2014very}, ResNet \cite{he2016deep}, DenseNet \cite{huang2017densely}, SqueezeNet \cite{iandola2016squeezenet} and Inception V3 \cite{szegedy2016rethinking}, and obtain 17 CNN based sketch recognition methods by using different layers of these CNN models. Our experimental results demonstrate that deeper CNN models can easily achieve superior performance over the handcrafted features. In particular, ResNet \cite{he2016deep} and DenseNet \cite{huang2017densely} outperform most of the existing methods. These results show that existing deep models can be noticeably effective architectures for sketch recognition.

We conduct extensive experiments on the TU-Berlin freehand sketch dataset \cite{eitz2012humans}. Our approach achieves remarkably higher performance than other state-of-the-art approaches. It is worth mentioning that the recognition accuracy of our approach is 11.27\% higher than the human performance. We also present detailed comparative results on the new Sketchy-R benchmark. Moreover, extra experimental results demonstrate the effectiveness of our approach to the task of sketch-based image retrieval.

The contributions of this paper are summarized as follows:
\begin{enumerate}
\item We present an automatic sketch-specific data augmentation (SSDA) method that relieves the cost of data collection for sketch recognition.
\item We introduce the Bezier pivot based deformation (BPD) to generate richer and more diverse training data.
\item We propose the mean stroke reconstruction (MSR) to create new sketches with smaller intra-class variances.
\item Extensive experiments are conducted on the TU-Berlin dataset, which indicates the proposed method outperforms all existing methods. In addition, we present an objective and comprehensive evaluation of different CNN models for sketch recognition.
\end{enumerate}

The remaining sections are organized as follows. We first briefly review the related work in freehand sketch recognition and data augmentation in Section \ref{sec:relatedwork}. We introduce the proposed Bezier pivot based deformation and mean stroke reconstruction methods in Section \ref{sec:methodology}. Experimental analysis and implementation details are provided in Section \ref{sec:experiments}. Finally, we articulate our conclusions in Section \ref{sec:conclusions}.

\section{Related Work}
\label{sec:relatedwork}
In this section, we first introduce some representative works in the field of freehand sketch recognition, which can be divided into two categories: handcrafted features based methods and deep learning based methods. Then we briefly present several closely related methods of data augmentation.

\subsection{Freehand Sketch Recognition}
Earlier works on sketch recognition like the Sketchpad \cite{sutherland1964sketchpad} and HUNCH \cite{herot1976graphical} systems have demonstrated the practical value of sketch recognition. However, this area is making slow progress due to the lack of sketch data. For example, the PaleoSketch system \cite{paulson2008paleosketch} can only recognize a few number of shapes such as circle and ellipse. To address this problem, Eitz et al. \cite{eitz2012humans} collect a large-scale freehand sketch dataset, which consists of 20,000 sketch images in 250 classes. After that, lots of outstanding works emerge on that dataset.

\textbf{Handcrafted features based methods.} The workflow of using handcrafted features for sketch recognition is almost the same as traditional object recognition in real images, which include feature extraction, representation, model training, and evaluation. One of biggest difference is that whether the feature or representation are specially designed for sketches. Furthermore, there also exists another kind of methods to make efforts at the later stage. Jayasumana et al. \cite{jayasumana2014optimizing} implement the kernel optimization on compact manifolds within the SVM framework. Li et al. \cite{li2015free} propose to fuse different types of features for sketch recognition by multi-kernel learning. Although the above-mentioned methods have made great achievements on the task of sketch recognition, the recognition performance still needs to be improved, just as A.Borji and L.Itti pointed in their paper \cite{borji2014human}.

\textbf{Deep learning based methods.} The deep neural networks significantly improve the performance of sketch recognition. For example, the Sketch-a-Net proposed by Yu et al. \cite{yu2015sketch} beats human at the recognition accuracy on TU-Berlin dataset, for the first time. It makes deep learning widely accepted by the researchers in the sketch related fields. Su et al. \cite{su2015multi} apply a multi-view CNN model for 3D shapes to recognize the 2D freehand sketches. Sarvadevabhatla et al. \cite{sarvadevabhatla2016enabling} introduce the recurrent neural network (RNN) to capture the temporal cues of sketch lines. But both methods only test on part of the TU-Berlin dataset, so it is difficult to evaluate their models fairly. Zhang et al. \cite{zhang2016sketchnet} propose a well-designed CNN architecture, which takes triplets of real images and sketches as the input. Yu et al. \cite{yu2017sketch} present a four-channel Siamese network and fuse its output by the joint Bayesian. However, these algorithms are too expensive when applied to large-scale datasets. Recent progress on the face sketch recognition proposes to take advantage of the adversarial sketch-photo transformation or deep local descriptor for better performance \cite{LiuLWPG18} \cite{YuHSDC19} \cite{PengWLG19}. In this paper, we demonstrate the superiority of some deeper CNN models to these elaborative networks. Therefore, we choose to explore other important problems for sketch recognition. With the solution of these problems, our approach achieves the state-of-the-art with significantly higher recognition accuracy on the TU-Berlin dataset.

\subsection{Data Augmentation}
The data augmentation technology plays a very important role in the area of machine learning and pattern recognition. As for image related fields, it can be classified into two categories: general and domain-specific methods.

\textbf{General augmentation methods.} The lack of training data makes it difficult to achieve a higher performance. In addition, it will increase the risk of overfitting at the training stage. Therefore, general augmentation methods are widely used in areas like image classification \cite{snoek2015scalable}, sketch-based image retrieval \cite{song2017deep}, sketch beautification \cite{Simo-SerraISI16}, and image super-resolution \cite{kim2016accurate}. These methods mainly include mirroring, random cropping, flipping, rotation, etc. In the experiments, we also implement the general technologies to augment the training sketch data.

\textbf{Domain-specific augmentation methods.} To further improve the recognition power of trained models, a huge number of specific methods are presented by exploring the intrinsic characters embedded in the corresponding domain. Especially when training data is difficult to collect or not publicly available, these methods seem more important. The existing methods are more focused on the area of face recognition \cite{tran2017regressing}, human pose recognition \cite{shotton2013real}, and object viewpoint estimation \cite{su2015render}. Obviously, all these methods are not suitable for the task of sketch recognition. The most related work is published by Yu et al. \cite{yu2017sketch}. They specifically design two kinds of data augmentation algorithm based on sketch removal and deformation, which can greatly enhance the diversity and scale of training sketch data. However, one of the requirements that must be met is to provide the temporal order of each stroke for all sketches, whereas our approach does not have this limitation.

\section{The Methodology}
\label{sec:methodology}
To address the problems of insufficient freehand sketches and huge intra-class variance, we propose a sketch-specific data argumentation (SSDA) method. Specifically, the proposed SSDA method consists of two novel approaches, namely Bezier pivot based sketch deformation and mean stroke reconstruction, which aim at improving the performance of sketch recognition from both improving sketch quality and increasing sketch quantity. Notice that, we employ the augmented datasets generated by SSDA for retraining the existing deep models as explained in Section \ref{sec:experiments}.

\subsection{Bezier Pivot based Sketch Deformation}
\label{sec:bezier}
As we have mentioned above, existing sketch-domain specific methods \cite{yu2017sketch} for data augmentation heavily rely on the temporal cues of sketch strokes, which greatly limits their applications. To solve this problem, we propose a Bezier pivot based deformation (BPD) approach to generate more diverse freehand sketches, which does not rely on any temporal information.

For a freehand sketch image $S$, we first convert it to a grayscale image and perform a simple threshold operation to obtain a binary image $B$. In our experiments, we set the threshold $t=128$. To extract the centerline $S ^\prime$, a morph based skeletonization method is applied to $B$. Specifically, it removes pixels on the boundaries of $B$ while preserves the Euler number. Then, we segment $S ^\prime$ into several disjoint square patches of size $a \times a$. We set $a=32$ for sketch images of size 256$\times$256. Considering that the lines of sketches have thickness, we extract the centerline before segmentation, which can avoid one short part of line being segmented into two patches. In each patch, we select the largest set of connected pixels as the main curve $T$, which is fitted by a cubic Bezier curve. The curves that only contain very few pixels are eliminated to ensure the fitting performance.

\begin{figure}[h]
    \centering
    \begin{minipage}[c]{0.45\linewidth}
    \centering
    \includegraphics[width=0.9\linewidth]{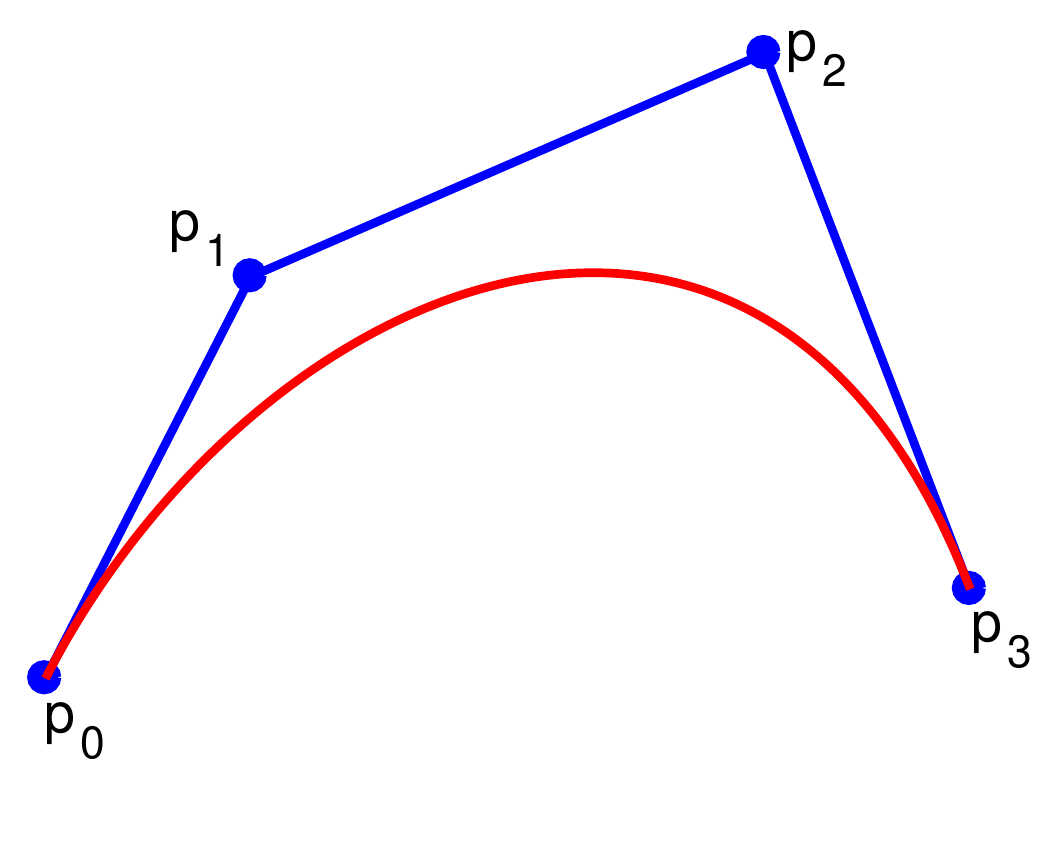}
    \end{minipage}
    \begin{minipage}[c]{0.45\linewidth}
    \centering
    \includegraphics[width=0.9\linewidth]{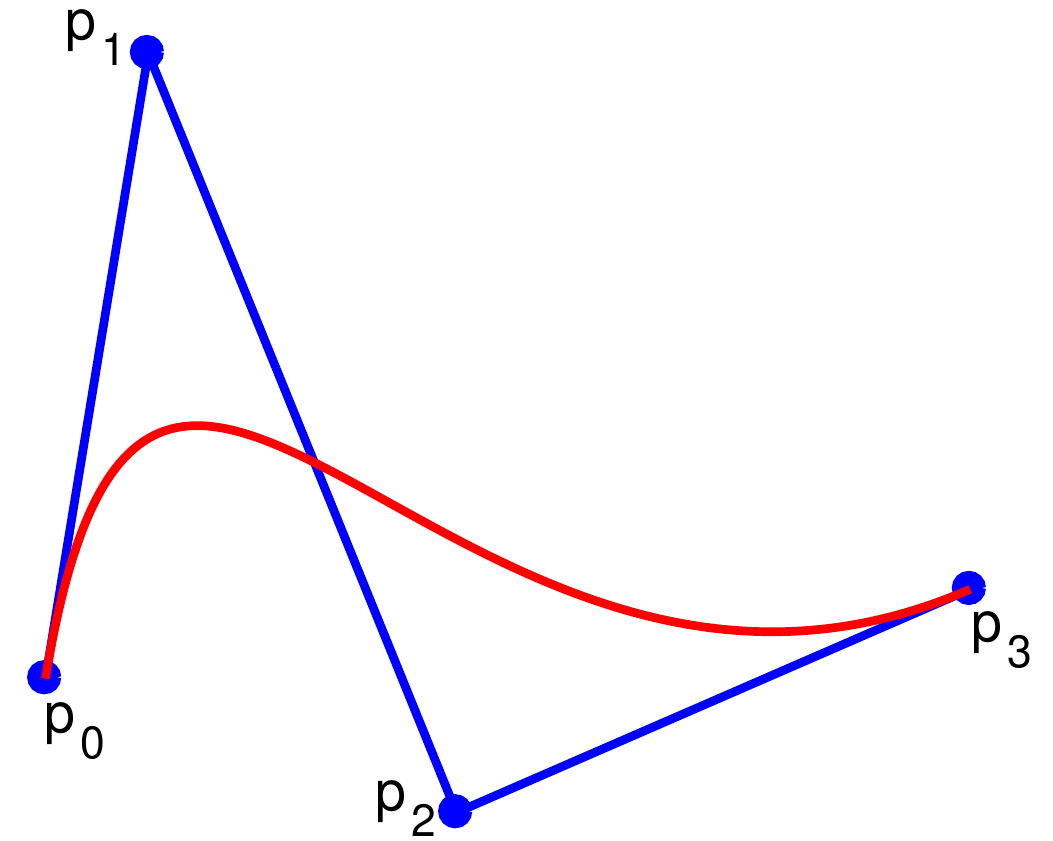}
    \end{minipage}
    \caption{Illustration of the cubic Bezier curves. $\textbf{p}_0$ and $\textbf{p}_3$ are the starting and ending points, while $\textbf{p}_1$ and $\textbf{p}_2$ are middle control pivots. With only four control pivots, it can represent a great diversity of curves.}
    \label{fig:bezier}
\end{figure}

The cubic Bezier curve can model a great diversity of curves by only four control pivots \cite{shaheen2017constrained}. As shown in Figure \ref{fig:bezier}, if we change the coordinates of the control pivots, we can get a series of points, which form a new curve. The function of a cubic Bezier curve is formulated as follows
\begin{equation}
	\begin{split}
        f = (1-t)^3 \textbf{p}_0 + 3t(1-t)^2 \textbf{p}_1 + 3t^2(1-t)\textbf{p}_2 + t^3 \textbf{p}_3
    \end{split}
\end{equation}
where $t \in [0, 1]$, $\textbf{p}_0$ and $\textbf{p}_3$ are the starting and ending points of the curve. The pivots $\textbf{p}_1$ and $\textbf{p}_2$ are the control points, which determine the curving shape. For a more concise representation, we introduce $\phi=1-t$ to shorten the formula and can obtain
\begin{equation}
	\begin{split}
        f = \phi^3 \textbf{p}_0 + 3t\phi^2 \textbf{p}_1 + 3t^2\phi\textbf{p}_2 + t^3 \textbf{p}_3
    \end{split}
    \label{eq:ft}
\end{equation}

Our goal is to generate more diverse sketches by deformation based on these Bezier control pivots of sketch patches. For the curve $T$ in each sketch patch, $\textbf{p}_0$ and $\textbf{p}_3$ can be obtained directly, while $\textbf{p}_1$ and $\textbf{p}_2$ are required to be computed. Supposing that the curve $T$ consists of $n$ points denoted by $v_i$ ($i=1, 2, \dots, n$), we propose to find the best-fitted Bezier curve by the Least Squares Method. The objective function is defined as
\begin{equation}
	\begin{split}
        f^{\ast} = \min L = \min \sum \limits_{i=1}^n (v_i - f(t_i))^2
    \end{split}
\end{equation}

We can get the curve function $f$ by minimizing the objective function, which can be solved by computing the partial derivatives of $L$ with respect to $\textbf{p}_1$ and $\textbf{p}_2$
\begin{equation}
	\begin{split}
        \frac{\partial L}{\partial \textbf{p}_1}  = 0
    \end{split}
    \label{eq:pd1}
\end{equation}
\begin{equation}
	\begin{split}
        \frac{\partial L}{\partial \textbf{p}_2}  = 0
    \end{split}
    \label{eq:pd2}
\end{equation}

By substituting Eq. (\ref{eq:ft}) into Eq. (\ref{eq:pd1}), we have
\begin{equation}
	\begin{split}
        &~~~~\frac{\partial \sum \limits_{i=1}^n (v_i - \phi_i^3 \textbf{p}_0 - 3t_i\phi_i^2 \textbf{p}_1 - 3t_i^2\phi_i\textbf{p}_2 - t_i^3 \textbf{p}_3)^2}{\partial \textbf{p}_1}\\
        &= \sum \limits_{i=1}^n 2(v_i - \phi_i^3 \textbf{p}_0 - 3t_i\phi_i^2 \textbf{p}_1 - 3t_i^2\phi_i\textbf{p}_2 - t_i^3 \textbf{p}_3) \times (-3 t_i\phi_i^2)\\
        &= \sum \limits_{i=1}^n 2(v_i - \phi_i^3 \textbf{p}_0 - t_i^3 \textbf{p}_3) \times (-3 t_i\phi_i^2)\\
        & \quad+\sum \limits_{i=1}^n 2(9t_i^2 \phi_i^4 \textbf{p}_1 + 9t_i^3 \phi_i^3 \textbf{p}_2)\\
        &= 0
    \end{split}
\end{equation}

From the above equation, we can easily obtain
\begin{equation}
	\begin{split}
    	\sum \limits_{i=1}^n 3t_i^2 \phi_i^4 \textbf{p}_1 + \sum \limits_{i=1}^n 3t_i^3 \phi_i^3 \textbf{p}_2 = \sum \limits_{i=1}^n t_i\phi_i^2(v_i - \phi_i^3 \textbf{p}_0 - t_i^3 \textbf{p}_3)\\
    \end{split}
    \label{eq:pd3}
\end{equation}

To simplify this formula, we introduce the notations $a_1$, $b_1$, and $\textbf{c}_1$ defined as follows to represent the coefficients of Eq. (\ref{eq:pd3}).
\begin{equation}
	\begin{split}
    	a_1 &= \sum \limits_{i=1}^n 3t_i^2 \phi_i^4\\
        b_1 &= \sum \limits_{i=1}^n 3t_i^3 \phi_i^3\\
        \textbf{c}_1 &= \sum \limits_{i=1}^n t_i\phi_i^2(v_i - \phi_i^3 \textbf{p}_0 - t_i^3 \textbf{p}_3)
    \end{split}
\end{equation}

Then Eq. (\ref{eq:pd3}) can be written as
\begin{equation}
	\begin{split}
    	a_1 \textbf{p}_1 + b_1 \textbf{p}_2 = \textbf{c}_1
    \end{split}
    \label{eq:pd4}
\end{equation}

Similarly, we can obtain the following equation from Eq. (\ref{eq:pd2})
\begin{equation}
	\begin{split}
        a_2 \textbf{p}_1 + b_2 \textbf{p}_2 = \textbf{c}_2
    \end{split}
    \label{eq:pd5}
\end{equation}
where the coefficients are defined as follows
\begin{equation}
	\begin{split}
    	a_2 &= b_1 = \sum \limits_{i=1}^n 3t_i^3 \phi_i^3\\
        b_2 &= \sum \limits_{i=1}^n 3t_i^4 \phi_i^2\\
        \textbf{c}_2 &= \sum \limits_{i=1}^n t_i^2 \phi_i(v_i - \phi_i^3 \textbf{p}_0 - t_i^3 \textbf{p}_3)
    \end{split}
\end{equation}

Through Eqs. (\ref{eq:pd4}) and (\ref{eq:pd5}), the variables $\textbf{p}_1$ and $\textbf{p}_2$ are expressed as
\begin{equation}
	\begin{split}
    	\textbf{p}_1 = \frac{b_2 \textbf{c}_1 - b_1 \textbf{c}_2}{a_1 b_2 - b_1 b_1}\\
        \textbf{p}_2 = \frac{a_1 \textbf{c}_2 - b_1 \textbf{c}_1}{a_1 b_2 - b_1 b_1}
    \end{split}
\end{equation}

\begin{figure*}[t]
    \centering
    \centerline{\includegraphics[width=1.0\linewidth]{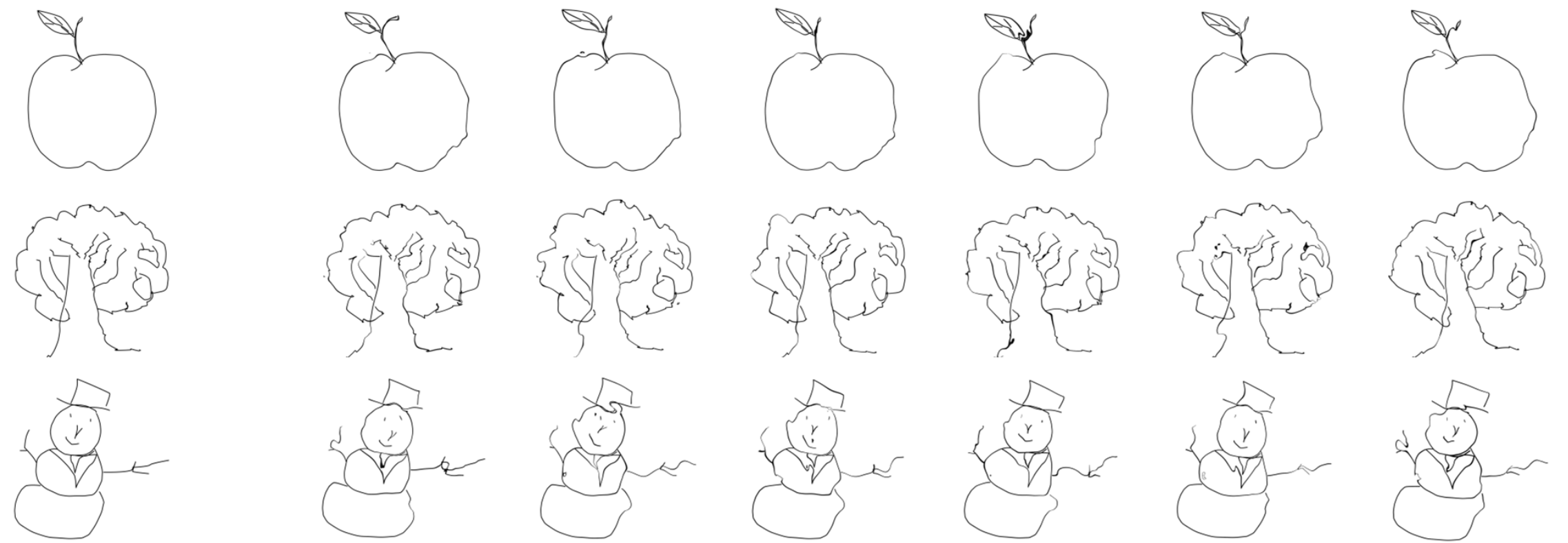}}
    \caption{Examples of sketches generated by the proposed BPD approach.The left one in each row is the original freehand sketch of the TU-Berlin dataset. The other 6 samples are deformed sketches generated by BPD.}
    \label{fig:deformation}
\end{figure*}

\begin{figure}[t]
    \centering
    \centerline{\includegraphics[width=0.65\linewidth]{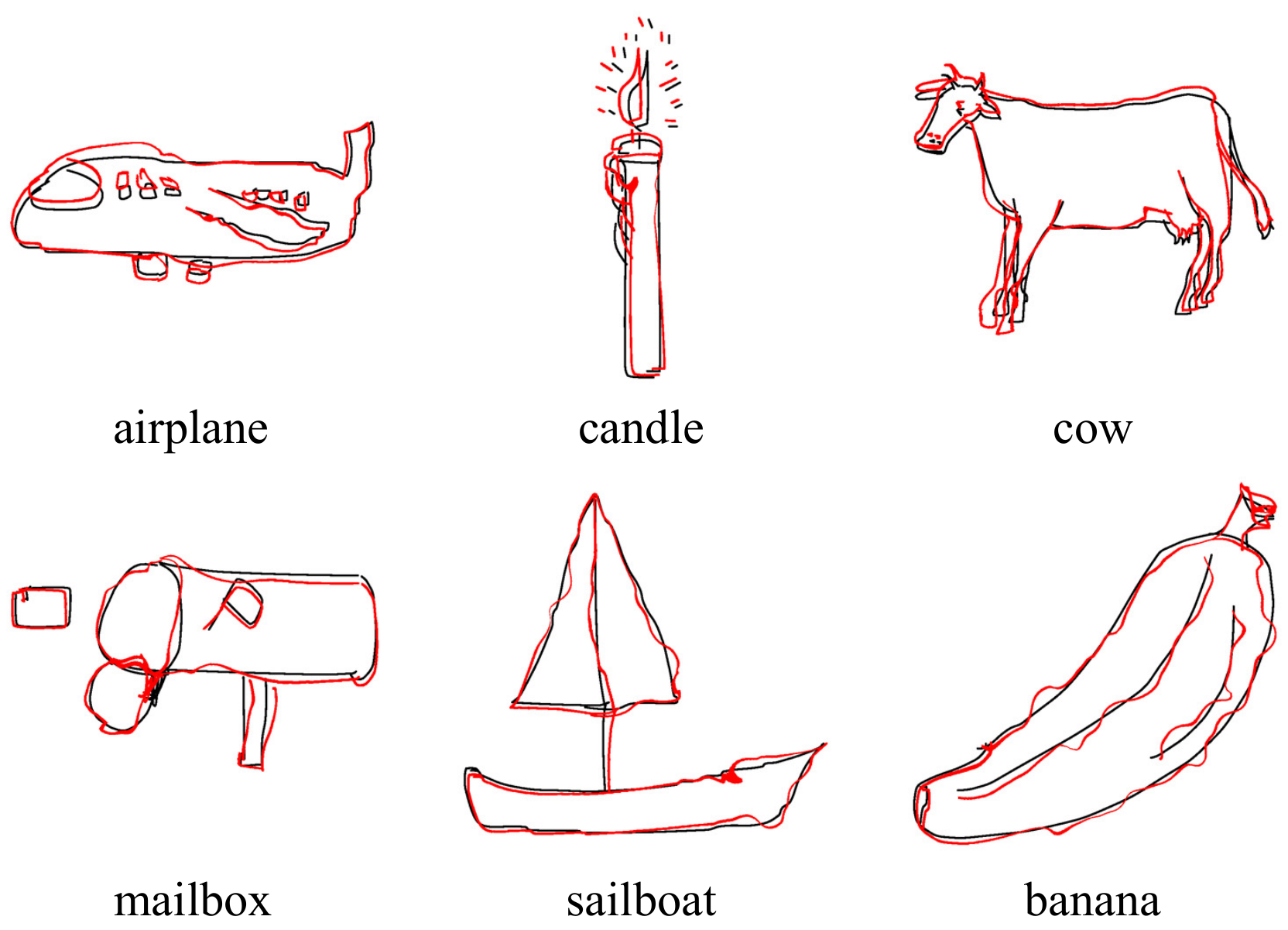}}
    \caption{Illustration of the deformation effect of our BPD approach. The sketches in black lines are original samples taken from the TU-Berlin dataset, while the sketches generated by BPD are shown in red lines.}
    \label{fig:reddeformation}
\end{figure}

Due to the highly abstract property of freehand sketches and the difference of drawing skill between humans, the drawn sketches show great diversity in many aspects, such as the curve length and bending degree. Considering the huge number of people and the differences in drawing skills, the variation of freehand sketches is closer to a stochastic process. Therefore, we apply a random shift $\Delta$ to the obtained control pivots $\textbf{p}$ to get the locations of new pivots $\textbf{p}^\prime$
\begin{equation}
	\begin{split}
    	\textbf{p}^\prime = \textbf{p} + \Delta
    \end{split}
\end{equation}
where $\Delta=(x, y)$, $x, y \in [-\alpha, \alpha]$, and $\alpha$ refers to the deformation degree. In our experiments, we set $\alpha=8$ for sketch images of size 256$\times$256. Based on these new control pivots, we perform the moving least squares algorithm \cite{schaefer2006image} to generate the deformed sketches. Figure \ref{fig:deformation} shows some examples of the generated sketches by the proposed BPD approach, from which we can see that our approach performs very well in generating more diverse sketches. To illustrate the deformation effect, we show some examples of new sketches generated by the proposed BPD approach in Figure \ref{fig:reddeformation}.

\subsection{Mean Stroke Reconstruction}
A freehand sketch is composed of several strokes, which are extremely diverse in length, thickness, radian, starting and ending points, etc. Even the simplest straight line shows great difference depending on the person drawing, skill, time cost, etc. It is the reason why the intra-class variance of freehand sketches is much bigger than real images, which makes the task of sketch recognition more challenging. If we can improve the quality of sketch images by reducing the intra-class variance, a model with higher recognition performance can be expected. Therefore, our goal is to generate new sketches with smaller intra-class variance on the original dataset. Without the need for a huge number of real images or other multi-source data, we aim to strengthen the classification power of models by improving the quality of sketches.

It is well known that discriminative patches in real images can be used as the mid-level visual representation \cite{singh2012unsupervised}. In freehand sketches, there also have the mean strokes, which can represent majority forms of sketch lines \cite{lim2013sketch}. Inspired by these works, we propose a novel approach of sketch generation based on mean stroke reconstruction (MSR). As the new sketches are constructed by the mean strokes, it have a smaller intra-class variance.

\begin{figure}[h]
    \centering
    \centerline{\includegraphics[width=0.65\linewidth]{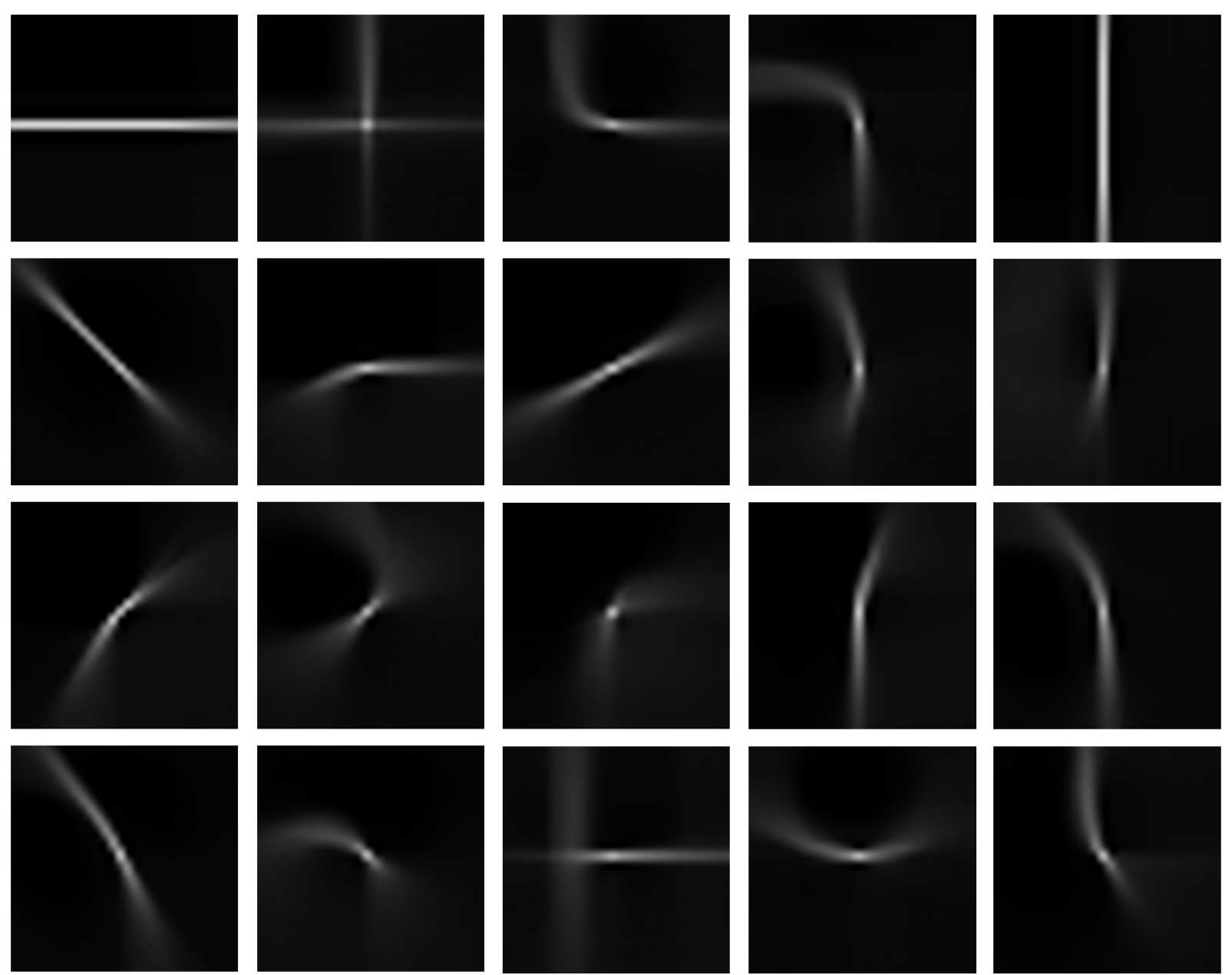}}
    \caption{Examples of mean strokes computed from freehand sketches. The obtained mean strokes are so rich that can represent a variety of basic sketch lines.}
    \label{fig:meanstroke}
\end{figure}

\textbf{Mean stroke computation.} Given the training sketch set, the first stage is to compute the mean strokes. We first resize a sketch $S$ to the size of 256$\times$256 and apply the same threshold and morph operation as Section \ref{sec:bezier} to get the skeleton $S ^\prime$. After that, taking each non-zero pixel $p_i$ as the center, the patch $s_i$ is extracted from $S ^\prime$ with a fixed size of 31$\times$31 pixels. Finally, tens of millions of patches are produced on the training sketch set. Such a huge number of patches bring great computational load to the subsequent algorithms. Therefore, we randomly select $1/\rho$ of patches and eliminate the others. The value of $\rho$ is determined jointly by the number of extracted patches and the size of available computing memory. For the TU-Berlin dataset, we set $\rho=3$ because a smaller $\rho$ will bring the problem of out of memory in our 32G memory computer. Similarly, we set $\rho=10$ for the Sketchy-R benchmark. For other databases, the setting of $\rho$ can be adjusted by comparing the number of training sketches in the TU-Berlin dataset and the memory size of us. HOG features \cite{dalal2005histograms} are extracted to describe all these remained patches, which are clustered by the k-means algorithm. Following the setup of \cite{lim2013sketch}, we set the cluster number $k=150$. Then, we can get the mean stroke $M_j$ by averaging all the sketch patches $s_{ij}$ belong to the cluster $j$, which is formulated as follows
\begin{equation}
	\begin{split}
        M_j = \sum \limits_{i=1}^\eta s_{ij} / \eta
    \end{split}
    \label{eq:meanstroke}
\end{equation}
where $\eta$ is the number of sketch patches in cluster $j$. The examples of the generated mean strokes are shown in Figure \ref{fig:meanstroke}, which shows that the obtained mean strokes include diverse line shapes.

\textbf{Sketch patch classification.} The second stage is to generate new sketches after obtaining the mean strokes. First, a classifier is learned to predict the labels of sketch patches. Because of the limitation of memory capacity, it is unrealistic to use all patches from the training set to train the classifier. Therefore, we randomly sample $m_1=100$ patches in each cluster and take the cluster id as its class label. In the experiments, we take the linear support vector machine (SVM) model as the patch classifier. Using a part of training patches inevitably weaken the classification power of SVM model. To address this problem, we propose to apply an ensemble method to get a more powerful classifier. Specifically, several SVM models are trained independently on the dataset of randomly sampled sketch patches. Then we perform the score-level fusion on the predicted scores output by these models. The class label with the highest score is appointed as the final prediction for the input sketch patch.

\begin{figure}[htbp]
    \centering
    \centerline{\includegraphics[width=0.75\linewidth]{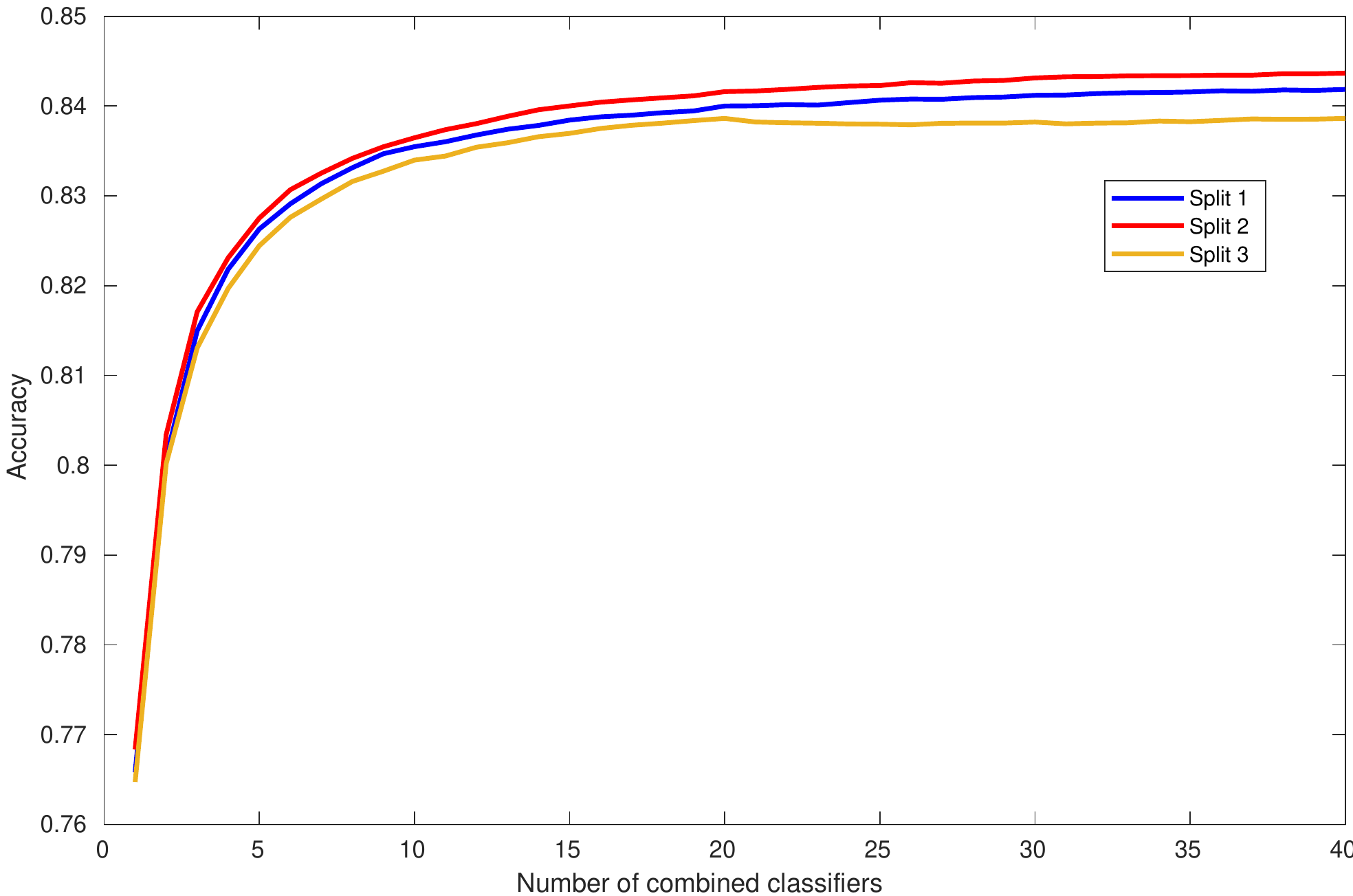}}
    \caption{The classification accuracies under different numbers of combined classifiers on three splits of the TU-Berlin dataset. The accuracy shows a sharp increment at first and gradually become more and more gentle.}
    \label{fig:classifier}
\end{figure}

\begin{figure}[htbp]
    \centering
    \centerline{\includegraphics[width=0.75\linewidth]{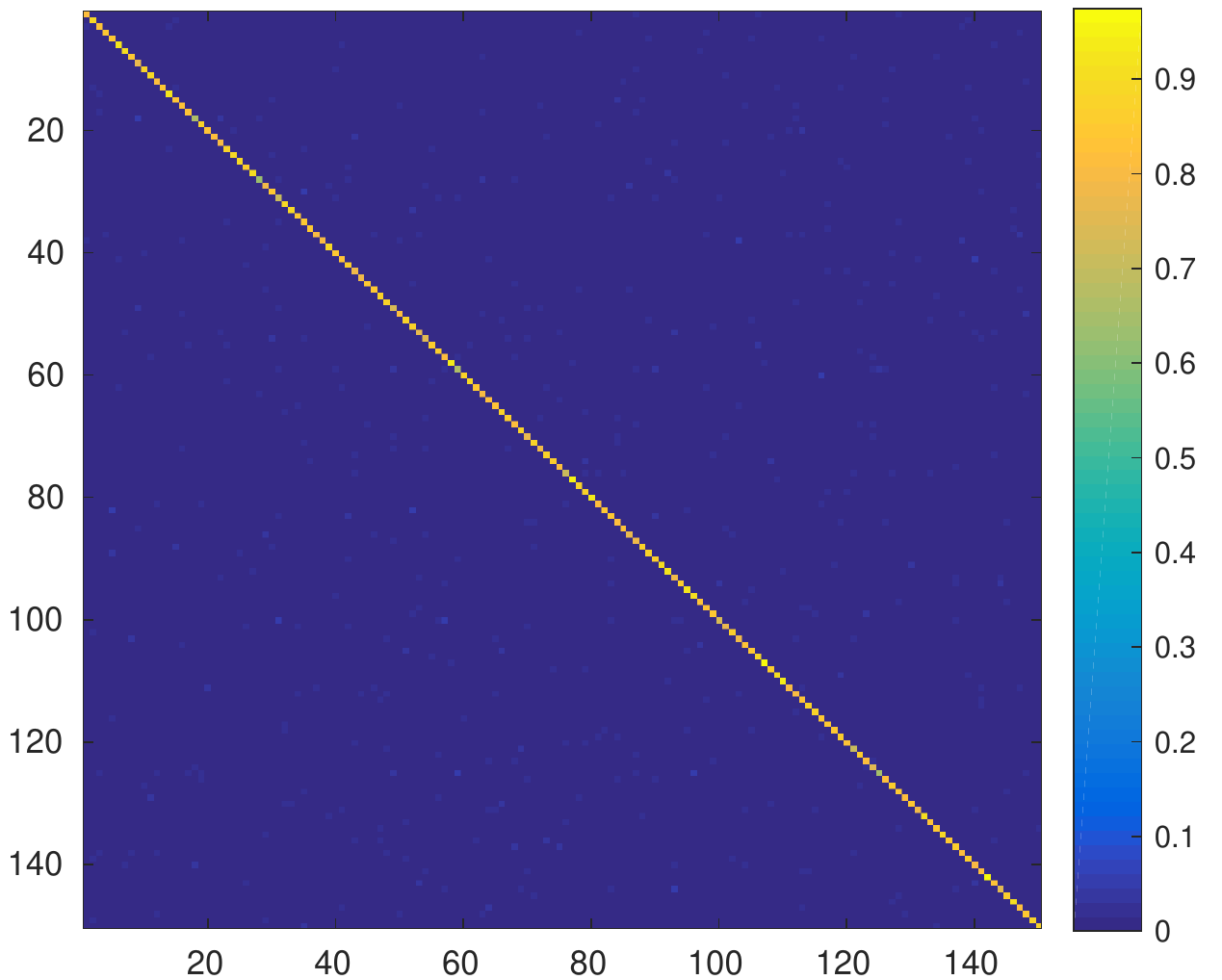}}
    \caption{The confusion matrix of the final patch classification model on the first split of the TU-Berlin dataset.}
    \label{fig:confusionmatrix}
\end{figure}

To find the most appropriate number $r$ of the combined SVM models, we evaluate the performance of classifiers under different $r$. The test data is collected by randomly sampling $m_2$ sketch patches from each cluster. The value of $m_2$ is determined by the cluster with the minimal number of patches. In the experiments, we set $m_2=7000$. The curves of classification performance on three splits of the TU-Berlin dataset \cite{eitz2012humans} are shown in Figure \ref{fig:classifier}. It can be seen that the performance presents a rising trend with the increase of $r$. Especially when there is a small number $r$, the performance increases drastically. As $r$ becomes larger and larger, the growth rate gradually comes to a standstill. Taking into account the performance and computation cost, we set $r=20$ in the experiments. The confusion matrix of the final patch classification model on the first split of the TU-Berlin dataset \cite{eitz2012humans} is shown in Figure \ref{fig:confusionmatrix}. It shows that the model performs very well among most of the clusters. Moreover, it also demonstrates that sketch patches belonging to the same cluster share a particular pattern, i.e. the mean stroke.

\begin{figure}[t]
    \centering
    \centerline{\includegraphics[width=0.75\linewidth]{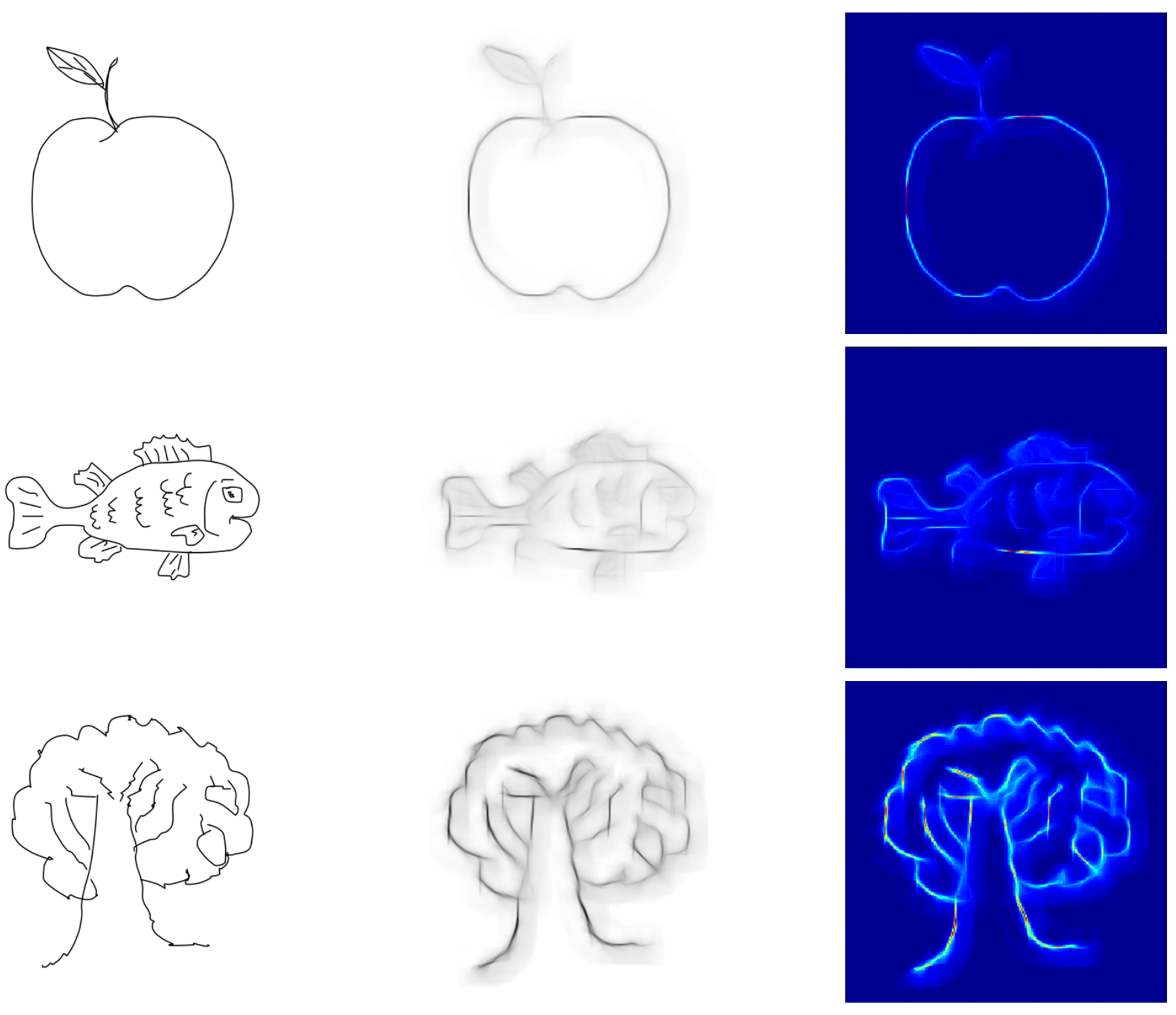}}
    \caption{Examples of the generated sketches by the proposed MSR method. Left: the original sketch, Middle: the generated sketch by MSR, Right: heat map of the middle sketch, in which the brighter part has a strong response to the mean stroke.}
    \label{fig:newsketch}
\end{figure}

\textbf{Sketch reconstruction.} The next problem is how to reconstruct the freehand sketches through the obtained mean strokes and classification model. To generate new sketches, we propose to replace the original sketch patches by weighted mean strokes. The extraction of patches for each freehand sketch is the same as we have mentioned above. For a sketch patch $s_i$, we first use the trained classifier to predict the cluster $j$ it belongs to. Then the sketch patch is replaced by a weighted mean stroke as follows
\begin{equation}
	\begin{split}
        s ^\prime _i = w_j \times M_j
    \end{split}
    \label{eq:newpatch}
\end{equation}
where $w_j$ is the weight of the mean stroke $M_j$. As the confusion matrix has revealed in Figure \ref{fig:confusionmatrix}, the classifier presents unbalanced performance on different classes. To reflect the probabilities to get the right predictions, $w_j$ is set as the normalized classification precision on class $j$. Finally, the new sketch $R$ is reconstructed by
\begin{equation}
	\begin{split}
        R = \frac{sum(s ^\prime)}{\sqrt{C}}
    \end{split}
    \label{eq:newsketch}
\end{equation}
where $sum(s ^\prime)$ means pixel-wise summation after each pixel of $s ^\prime$ maps to the location in the original sketch image. The $C$ is a matrix in which $C(p, q)$ counts the number of pixels mapped to the location $(p, q)$, while $\sqrt{C}$ is taking the square root of each element. The division operation is also conducted in element-wise. Figure \ref{fig:newsketch} illustrates some examples of the generated sketches by the proposed MSR method.

\section{Experiments}
\label{sec:experiments}
In this section, we first introduce the datasets used for evaluation and describe the implementation details. Second, we conduct a comprehensive comparison of different types of CNN models in the task of sketch recognition. To achieve a more comprehensive understanding of the proposed approach, we evaluate the contributions of each component and present an ablation study via extensive experiments. After that, we compare the classification performance of our approach with several state-of-the-art methods. Furthermore, we introduce a new benchmark for sketch recognition. Finally, we supplement some experiments to further demonstrate the practical value of our approach for the task of sketch-based image retrieval.

\subsection{Datasets and Experimental Settings}
The TU-Berlin dataset\footnote{\url{http://cybertron.cg.tu-berlin.de/eitz/projects/classifysketch}} \cite{eitz2012humans} has 20,000 freehand sketches collected by Amazon Mechanical Turk (AMT). All sketches are equally distributed in 250 object classes, i.e. each class has 80 sketches. After constructing the dataset, the authors conduct a human classification experiment. The result shows that human can only correctly recognize 73.1\% of sketches, which demonstrates that freehand sketch recognition is a very challenging task. Following the existing works \cite{yu2017sketch} \cite{yu2015sketch}, we evaluate the proposed approach by three-fold cross-validation. That is, we have three splits in total on the dataset, where each split takes two folds for training and the rest one for test.

Most of the existing works on sketch recognition are conducted on the TU-Berlin dataset. To further facilitate future research, we introduce the public Sketchy dataset\footnote{\url{http://sketchy.eye.gatech.edu/}} \cite{sangkloy2016sketchy} as an additional benchmark for evaluation. The Sketchy dataset is published for the task of sketch-based image retrieval, which consists of 75,471 sketch images unevenly distributed in 125 object classes. Among the 125 classes, there are 100 categories that also exist in the TU-Berlin dataset. Because of the mistake in the process of human drawing, there are 918 sketches marked as erroneous. We abandon these completely wrong samples and save the other 74,553 sketches to construct the Sketchy-R benchmark. Same as the TU-Berlin dataset, three-fold cross-validation is performed on the Sketchy-R benchmark.

\subsection{Implementation Details}
When implementing the proposed MSR approach, we use the source code of LIBSVM \cite{chang2011libsvm} released on their website\footnote{\url{http://www.csie.ntu.edu.tw/~cjlin/libsvm}}. Following \cite{PengWWQ16}, we apply the one-vs-all strategy to train models and fix the parameter $c=149$ as the class number $k=150$. For the HOG features, we set the cell size to [8, 8] and the dimension of final feature vectors for sketch patches is 144. In the experiments, we use the trained models and obtained mean strokes to reconstruct every sketches in the training and test dataset for each split independently.

We implement the CNN models using a publicly available deep learning framework named PyTorch. In all experiments, we set the initial learning rate to 0.001 and decrease it by a factor of 10 every 7 epochs. The training process is terminated after 25 epochs. We adopt the cross-entropy loss and stochastic gradient descent (SGD) with 0.9 momentum in the training stage. The batch size is set to 20 for all models unless otherwise indicated. During training, sub-images of 224$\times$224\footnote{The Inception V3 model \cite{szegedy2016rethinking} is an exception, which takes images of 299$\times$299 as the input.} are randomly cropped from the input sketches and the random horizontal flip is performed. We shuffle all training data in each epoch. In the test stage, only the center crop is conducted.

\subsection{Comparative Results of Different CNN Models}
To explore the performance of different kinds of CNN models for freehand sketch recognition, we conduct a comprehensive experiment on three splits of the TU-Berlin dataset. We select 6 widely used CNN architectures for the evaluation, including AlexNet \cite{krizhevsky2012imagenet}, VGG \cite{simonyan2014very}, ResNet \cite{he2016deep}, DenseNet \cite{huang2017densely}, SqueezeNet \cite{iandola2016squeezenet} and Inception \cite{szegedy2016rethinking}. According to the different number of layers, we totally get 17 CNN models. To evaluate their performance, we transfer the CNN models pre-trained on Imagenet \cite{deng2009imagenet} to the task of sketch recognition by fine-tuning. In particular, the batch size of SqueezeNet is not the same as other models in the experiments. When the batch size is set to 20, its performance is very unstable. After many times of trials, we find that the batch size of 8 is a better choice for SqueezeNet.
\begin{table}[t]
\centering
\caption{Comparison of different CNN models on the TU-Berlin dataset.}
\begin{tabular}{lcccc}
\toprule
{} & {~~Split1~~}  & {~~Split2~~}  & {~~Split3~~}  & {~~Average~~}\\
\midrule
{SqueezeNet1.0 \cite{iandola2016squeezenet}} & 61.32\%  & 54.06\%  & 60.30\% & 58.56\%\\
{SqueezeNet1.1 \cite{iandola2016squeezenet}} & 63.38\%  & 59.59\%  & 63.82\%  & 62.26\%\\
{AlexNet \cite{krizhevsky2012imagenet}} & 68.63\%  & 68.61\%  & 69.48\%  & 68.91\%\\
{Inception V3 \cite{szegedy2016rethinking}} & 74.45\%  & 75.69\%  & 75.08\%  & 75.07\%\\
{VGG-11 \cite{simonyan2014very}} & 74.31\%  & 72.86\%  & 72.95\%  & 73.37\%\\
{VGG-13 \cite{simonyan2014very}} & 75.22\%  & 72.55\%  & 73.35\%  & 73.71\%\\
{VGG-16 \cite{simonyan2014very}} & 75.17\%  & 74.62\%  & 74.25\%  & 74.68\%\\
{VGG-19 \cite{simonyan2014very}} & 76.42\%  & 74.92\%  & 75.97\%  & 75.77\%\\
{ResNet-18 \cite{he2016deep}} & 75.40\%  & 73.16\%  & 73.24\%  & 73.93\%\\
{ResNet-34 \cite{he2016deep}} & 76.58\%  & 76.76\%  & 76.95\%  & 76.76\%\\
{ResNet-50 \cite{he2016deep}} & 76.92\%  & 76.76\%  & 77.48\%  & 77.05\%\\
{ResNet-101 \cite{he2016deep}} & 78.09\%  & 78.83\%  & 79.59\%  & 78.84\%\\
{ResNet-152 \cite{he2016deep}} & \textbf{79.25\%}  & \textbf{79.79\%}  & \textbf{80.03\%}  & \textbf{79.69\%}\\
{DenseNet-121 \cite{huang2017densely}} & 77.23\%  & 76.74\%  & 76.19\%  & 76.72\%\\
{DenseNet-169 \cite{huang2017densely}} & 78.42\%  & 77.97\%  & 78.80\%  & 78.40\%\\
{DenseNet-201 \cite{huang2017densely}} & 79.05\%  & 78.50\%  & 79.11\%  & 78.89\%\\
{DenseNet-161 \cite{huang2017densely}} & \textbf{79.85\%}  & \textbf{79.32\%}  & \textbf{79.48\%}  & \textbf{79.55\%}\\
\bottomrule
\end{tabular}
\label{tab:cnns}
\end{table}

The comparative results are shown in Table \ref{tab:cnns}. It can be seen that: 1) simple networks like SqueezeNet and AlexNet get the worst recognition performance which is far beneath human. 2) The VGG-19 and Inception V3 show moderate performance, which are slightly better than human but still cannot compare with the state-of-the-art methods such as SN2.0 \cite{yu2017sketch}. 3) The ResNet-152 and DenseNet-161 achieve the best performance among these CNN models, which are already better than SN2.0 \cite{yu2017sketch}. It should be noted again that the models are directly obtained by fine-tuning the pre-trained models on real images. In view of these observations, we conclude that deeper CNN models like ResNet and DenseNet can be noticeably effective architectures for sketch recognition. Therefore, we select ResNet and DenseNet as the base models for evaluation.

\subsection{Ablation Study}
To evaluate the contributions of the proposed Bezier pivot based deformation (BPD) approach, we test the classification accuracy of applying BPD alone on the TU-Berlin dataset. Actually, the BPD approach can generate countless sketches, which is unpractical on limited computation resources. In the experiments, we use BPD to generate 10 new sketches for each sketch of the training set. Together with the original sketches, we finally obtain 11 times training data. The accuracies and improvements compared to the original models are reported in Tabel \ref{tab:bpd}. It shows that the proposed BPD approach achieves better performance than the original models (Ori) on all three splits. After implementing BPD to generate more diverse freehand sketches for the model training, we get 2.40\% and 2.28\% performance improvements on average over ResNet-152 and DenseNet-161, respectively. It demonstrates that the proposed BPD approach is very effective for sketch recognition.
\begin{table}[h]
\centering
\caption{Evaluation on the contributions of Bezier pivot based deformation (BPD).}
\begin{tabular}{lccc}
\toprule
{} & {~~Ori~~}  & {~~BPD~~} & {Improvement}\\
\midrule
{ResNet-152} & 79.69\%  & \textbf{82.09\%}  & +2.40\%\\
{DenseNet-161} & 79.55\%  & \textbf{81.83\%}  & +2.28\%\\
\bottomrule
\end{tabular}
\label{tab:bpd}
\end{table}

%\subsection{Contributions of MSR}
We evaluate the contributions of the proposed mean stroke reconstruction (MSR) in the same way as BPD. From Table \ref{tab:msr}, we can see that performing our MSR approach alone can obtain equivalent performance on the original models. The new sketches generated by MSR reduce the intra-class variance, while at the same time losing some individual information to a certain extent. Therefore, we propose to combine the original models with MSR by score fusion. That is to say, we directly add the output scores of two models together and take the class label with the highest score as the final prediction. As shown in Table \ref{tab:msr}, the accuracies of ResNet-152 and DenseNet-161 are improved by an average of 1.35\% and 1.08\%, respectively. These results demonstrate that the proposed MSR approach plays a complementary role to the existing CNN models.
\begin{table}[h]
\centering
\caption{Evaluation on the contributions of mean stroke reconstruction (MSR).}
\begin{tabular}{lcccc}
\toprule
{} & {Ori}  & {MSR} & {Fusion} & {Improvement}\\
\midrule
{ResNet-152} & 79.69\%  & 79.65\%  & \textbf{81.04\%}  & +1.35\%\\
{DenseNet-161} & 79.55\%  & 79.92\%  & \textbf{80.63\%}  & +1.08\%\\
\bottomrule
\end{tabular}
\label{tab:msr}
\end{table}

%\subsection{Combination of MSR and BPD}
The proposed BPD approach can be considered to improve the classification performance by augmenting the dataset size, while the MSR aims to fulfill this goal by improving the data quality. They are complementary to each other. Therefore, we combine the two approaches together for freehand sketch recognition. Here, we also adopt the score fusion which is very simple and effective. The evaluation results are shown in Table \ref{tab:msrbpd}. Compared to the original models, the combination of MSR and BPD achieves 3.70\% and 3.68\% higher classification accuracies on average. Especially for the ResNet-152 model, the combination of MSR and BPD brings a surprising 4.24\% performance improvement on the first split of the TU-Berlin dataset. All these results have demonstrated the effectiveness of our approach.

\begin{table}[h]
\centering
\caption{Evaluation on the contributions of MSR + BPD.}
\begin{tabular}{lccc}
\toprule
{} & {~~Ori~~}  & {~~MSR+BPD~~} & {Improvement}\\
\midrule
{ResNet-152} & 79.69\%  & \textbf{83.39\%}  & +3.70\%\\
{DenseNet-161} & 79.55\%  & \textbf{83.23\%}  & +3.68\%\\
\bottomrule
\end{tabular}
\label{tab:msrbpd}
\end{table}

%\subsection{Fusion of Different CNN Models}
It is generally known that different types of features or models capture different kinds of particular characteristics. Therefore, many researchers propose to combine different features or models together to achieve a higher performance \cite{qi2019hedging} \cite{zhang2015adaptive} \cite{zhao2017continuous}. In this paper, we integrate the two CNN models of ResNet-152 and DenseNet-161 to further improve the performance. Extensive results are reported in Table \ref{tab:fusion}, where the fused MSR+BPD refers to our full model (SSDA). Once again, the results prove that the fusion of different CNN models can produce a higher accuracy. Most importantly, our approach achieves a new state-of-the-art with a remarkable classification accuracy of 84.27\% on the TU-Berlin dataset.
\begin{table}[h]
\centering
\caption{Comparison of different components after fusing two CNN models of ResNet-152 and Densenet-161.}
\begin{tabular}{ccccc}
\toprule
{Ori} & {MSR} & {BPD} & {MSR+Ori} & {MSR+BPD}\\
\midrule
81.04\% & 80.88\% & 83.38\% & 82.13\% & \textbf{84.27\%}\\
\bottomrule
\end{tabular}
\label{tab:fusion}
\end{table}

\begin{figure}[t]
    \centering
    \centerline{\includegraphics[width=0.75\linewidth]{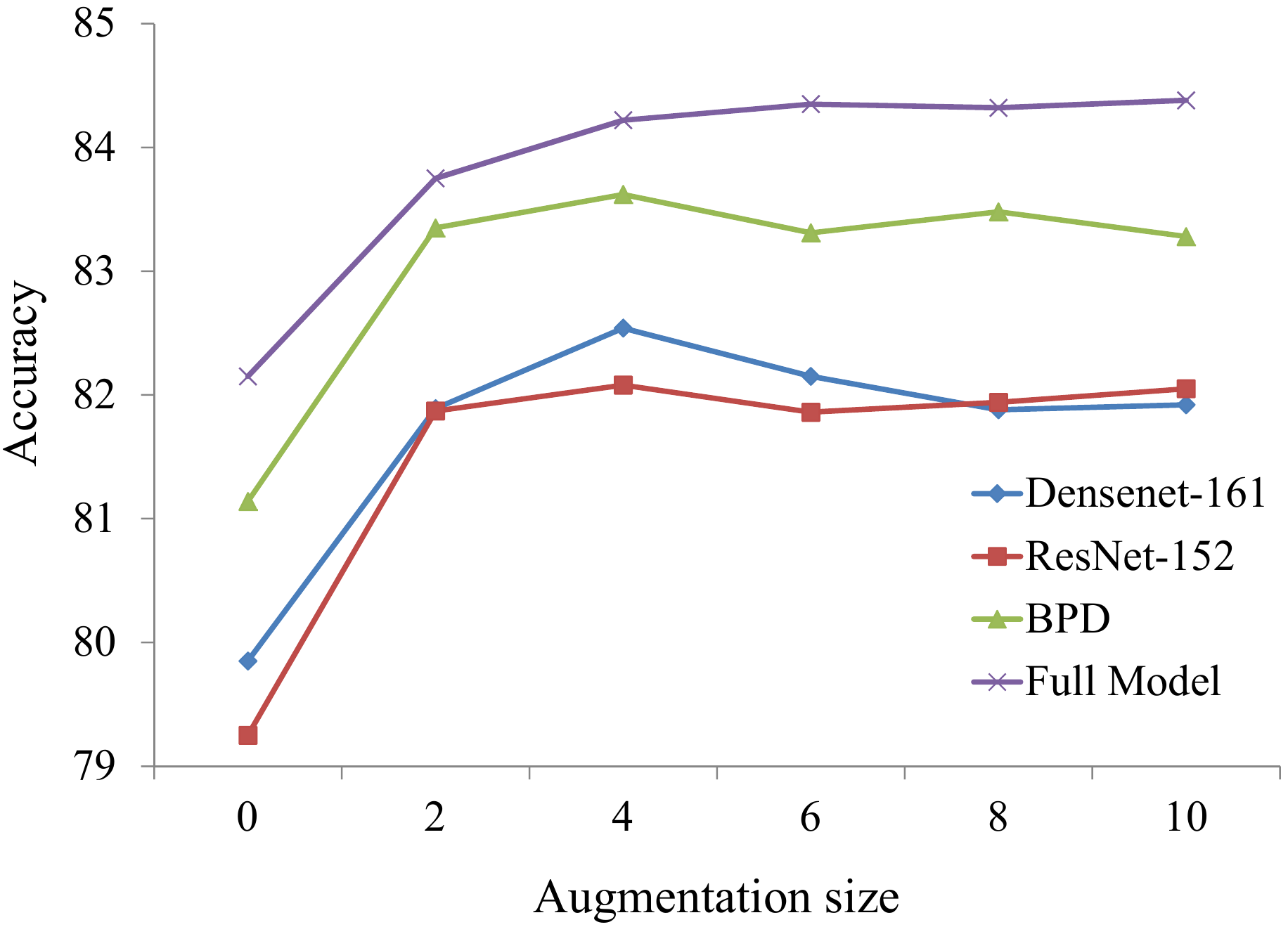}}
    \caption{Impact of augmentation size for the classification performance. `BPD' means the fused model of `ResNet-152' and `DenseNet-161', while `Full Model' combines it with the `MSR' approach. `0' refers to the original training set without data augmentation. `2, 4, 6, 8, 10' are the number of new sketches generated by our BPD approach on the training set. }
    \label{fig:augmentationsize}
\end{figure}

%\subsection{Impact of Augmentation Size}
When implementing the proposed BPD approach, the augmentation size is an important factor. To find the most appropriate number of new sketches generated by BPD for model training, we evaluate the classification performance under different augmentation sizes. As shown in Figure \ref{fig:augmentationsize}, the performance of these models dramatically increases when the augmentation size is small. As it increases from 2 to 6, there are some fluctuations for ResNet-152, DenseNet-161, and BPD, while our full model maintains a gentle rise. With the size increasing to 10, it eventually reaches a high performance and becomes stable. Therefore, to take performance and computation cost into account, we set the augmentation size to 10 in all experiments.

To further demonstrate the effectiveness of the proposed approach, we make a comparison with four types of data augmentation methods on the split1 of the TU-Berlin dataset. 1) Same as \cite{Sarvadevabhatla17}, we perform different degrees of rotations (0, $\pm$10, $\pm$20, $\pm$30) and mirroring on the original sketch image, which finally outputs 14 augmented images for each sketch. 2) We apply diverse degrees of rotation (0, $\pm$5, $\pm$10, $\pm$15, $\pm$20, $\pm$25, $\pm$30) to generate 13 times the size of the training data. 3) Following {\cite{Graham14}}, we implement a randomized mix of translations, rotations, stretching and shearing operations to generate 10 deformed sketches. 4) We use the stroke removal, local and global deformation proposed in {\cite{yu2017sketch}} to output 10 new sketches. The above methods can enlarge the training dataset to the same granularity as our approach. The comparative results are shown in Table \ref{tab:normalaug}. We can see that these augmentation methods bring slight performance improvement to the original model, which is significantly lower than our SSDA approach. The results demonstrate the superior performance of the proposed sketch-specific data augmentation (SSDA) approach compared to existing data augmentation methods.

\begin{table}[h]
\footnotesize
\centering
\caption{Comparison with other data augmentation methods on the TU-Berlin dataset.}
\begin{tabular}{lccccccc}
\hline
{} & {Ori} & {rotation} & {rot+mir} & {\cite{Graham14}} & {\cite{yu2017sketch}} & {BPD} & {our SSDA}\\
\hline
{ResNet-152} & 79.25\% & 80.66\% & 80.55\% & 80.46\% & 80.88\% & 82.05\% & \textbf{83.49\%}\\
{DenseNet-161} & 79.85\% & 80.67\% & 80.02\% & 80.58\% & 80.00\% & 81.92\% & \textbf{83.25\%}\\
\hline
\end{tabular}
\label{tab:normalaug}
\end{table}

\subsection{Comparison with State-of-the-Art Methods}
We compare the proposed SSDA approach with several state-of-the-art methods for freehand sketch recognition on the TU-Berlin dataset. The compared methods include traditional handcrafted features based algorithms and CNN based methods. The results are shown in Table \ref{tab:comp}, from which we can observe that: 1) without introducing any external data, our approach achieves state-of-the-art performance which shows a clear advantage to existing methods. 2) Our approach beats human on the task of sketch recognition by a remarkable 11.17\% increase. 3) Compared to the handcrafted features based algorithms, CNN based methods can easily gain better performance. 4) The accuracy of our approach is 6.32\% higher than SN2.0 \cite{yu2017sketch} which is the most representative method for sketch recognition. Considering that SN2.0 \cite{yu2017sketch} has an elaborately designed complex structure and relies on the temporal cues of sketch lines, the performance improvement achieved by our approach seems even more significant.

\begin{table}[h]
\footnotesize
\centering
\caption{Comparison of the recognition performance with state-of-the-art methods on the TU-Berlin dataset.}
\begin{tabular}{lc}
\hline
{} & {~~Accuracy~~}\\
\hline
{HOG-SVM \cite{eitz2012humans}} & 56\%\\
{MKL-SVM \cite{li2015free}} & 65.81\%\\
{FV-SP \cite{schneider2014sketch}} & 68.9\%\\
{AlexNet \cite{krizhevsky2012imagenet}} & 68.91\%\\
{SN1.0 \cite{yu2015sketch}} & 74.9\%\\
{Inception V3 \cite{szegedy2016rethinking}} & 75.07\%\\
{VGG-19 \cite{simonyan2014very}} & 75.77\%\\
{Zhou et al. \cite{ZhouJ20}} & 76\%\\
{SN2.0 \cite{yu2017sketch}} & 77.95\%\\
{Hybrid CNN \cite{ZhangHZPZW20}} & 78\%\\
{Zhang et al. \cite{ZhangSLGCF19}} & 82.95\%\\
\hline
{\textbf{Our SSDA}} & \textbf{84.27\%}\\
{Human} & 73.1\%\\
\hline
\end{tabular}
\label{tab:comp}
\end{table}

\subsection{Classification Results on the Sketchy-R Benchmark}
Same as the TU-Berlin dataset, we select ResNet-152 and DenseNet-161 as baselines. The classification results of our approach, these two models, and the combination of them on three splits of the Sketchy-R benchmark are presented in Table \ref{tab:sketchy}. The results show that our approach achieves the best performance. We can see that the classification accuracies of ResNet-152 and DenseNet-161 on the Sketchy-R benchmark are far higher than the TU-Berlin dataset. There are two reasons contributed to this difference: 1) the size of the Sketchy-R benchmark is much bigger than the TU-Berlin dataset. Specifically, the former has an average of 596 sketch images for each category, which is 7.45 times of the latter. 2) All sketches in the Sketchy-R benchmark are drawn according to the objects of real images while only a random category name is given for the TU-Berlin dataset. It leads that the sketches of the TU-Berlin dataset are more abstract than the Sketchy-R benchmark, which makes the TU-Berlin dataset more challenging. We hope the Sketchy-R benchmark can provide some help to the future research and application of sketch recognition.
\begin{table}[h]
\centering
\caption{Classification results on the Sketchy-R benchmark. }
\begin{tabular}{cccc}
\toprule
{ResNet-152 \cite{he2016deep}} & {DenseNet-161 \cite{huang2017densely}} & {Combination} & {\textbf{Our SSDA}}\\
\midrule
92.86\% & 92.49\% & 93.75\% & \textbf{95.57\%}\\
\bottomrule
\end{tabular}
\label{tab:sketchy}
\end{table}

\vspace{-10pt}

\subsection{Further Applications for Sketch-Based Image Retrieval}
Sketch-based image retrieval (SBIR) is strongly related to the task of sketch recognition as they can usually share the base networks. To demonstrate the practical value of the proposed approach, we integrate our approach into the training pipeline of existing SBIR networks and evaluate the performance on the QMUL FG-SBIR datasets \cite{song2017deep} \cite{yu2016sketch}.

We select two outstanding methods named Triplet SN \cite{yu2016sketch} and DSSA \cite{song2017deep} as our baseline models. Considered that both methods take triplets as the input of their networks, we apply the proposed BPD approach to create new training sketches as the anchor samples, which finally generate 10 times more triplets for the model training. Following the works of Triplet SN \cite{yu2016sketch} and DSSA \cite{song2017deep}, we use the same experimental settings and take the top K accuracy (acc.@K) as the evaluation metric. The comparative results against baselines on the QMUL FG-SBIR dataset (acc.@1) are shown in Table \ref{tab:sbir}. We can see that there are significant performance improvements for both baseline networks when integrated with the proposed BPD approach. The experimental results demonstrate the effectiveness of our approach for fine-grained instance-level SBIR.
\begin{table}[h]
\centering
\caption{Comparative results against baselines on the QMUL FG-SBIR dataset (acc.@1).}
\begin{tabular}{clccc}
\hline
{} & {} & {~~Ori~~}  & {~~BPD~~} & {Improvement}\\
\hline
\multirow{2}{*}{Shoe} & {Triplet SN \cite{yu2016sketch}} & 52.17\%  & \textbf{56.52\%}  & +4.35\%\\
                {} & {DSSA \cite{song2017deep}} & 58.26\%  & \textbf{61.74\%}  & +3.48\%\\
\hline
\multirow{2}{*}{Chair} & {Triplet SN \cite{yu2016sketch}} & 72.16\%  & \textbf{78.35\%}  & +6.19\%\\
                {} & {DSSA \cite{song2017deep}} & 79.38\%  & \textbf{80.41\%}  & +1.03\%\\
\hline
\multirow{2}{*}{Handbag} & {Triplet SN \cite{yu2016sketch}} & 39.88\%  & \textbf{43.45\%}  & +3.57\%\\
                {} & {DSSA \cite{song2017deep}} & 48.21\%  & \textbf{49.40\%}  & +1.19\%\\
\hline
\end{tabular}
\label{tab:sbir}
\end{table}

\vspace{-10pt}

\subsection{Qualitative Results}
\begin{figure}[h]
    \centering
    \centerline{\includegraphics[width=0.75\linewidth]{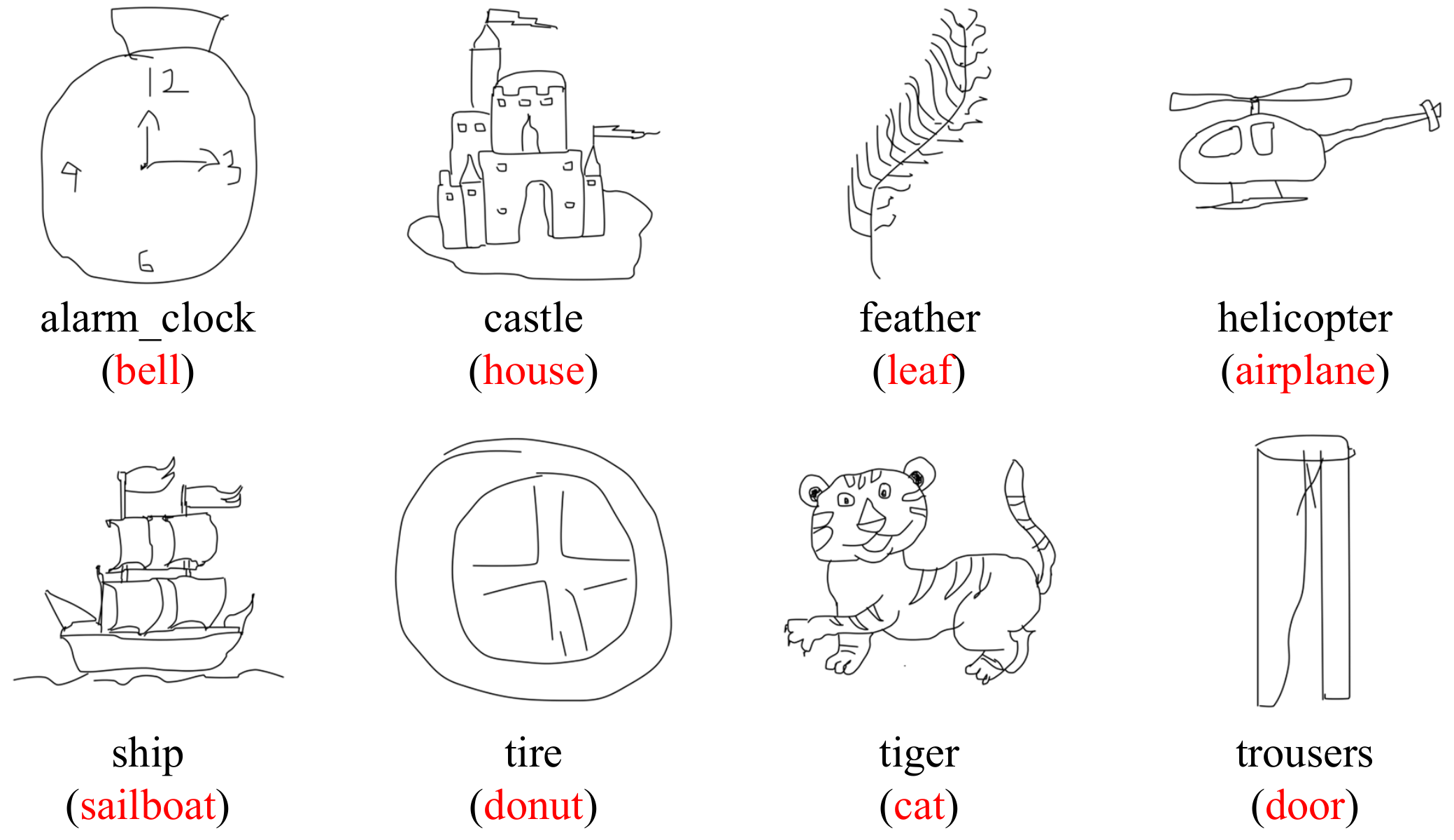}}
    \caption{Illustration of freehand sketches misclassified by human while our approach correctly recognize. The true class label of each category is shown in black below the sketch images, while the red word in brackets presents the wrong predicted label output by human.}
    \label{fig:humanflase}
\end{figure}

We show some examples of sketches misclassified by human while our approach recognizes them successfully. As illustrated in Figure \ref{fig:humanflase}, our approach can recognize lots of tough examples that are from two analogous classes or have the similar appearance. It makes our approach can beat human with significant higher performance.

There are two key reasons for the low performance of human on the task of freehand sketch recognition: 1) unlike the object recognition of real images, the training samples of freehand sketches are very limited for human. Some people have never seen any examples for several categories of sketches. Human more depend on the accumulated experience from the real world, while the CNN based methods heavily rely on a huge size of training data. 2) To some subtle differences between sketches, human is not as sensitive as the CNN models. Thus, human often make mistakes when faced with similar sketches.

\section{Conclusions}
\label{sec:conclusions}
In this paper, we investigate sketch-specific data augmentation methods to address the problems of lacking training data and the huge intra-class variance in freehand sketch recognition. To address the first problem, we introduce a Bezier pivot based deformation (BPD) method to create more diverse sketches. For the second problem, we propose a mean stroke reconstruction (MSR) based approach to generate new types of sketches with a smaller intra-class variance. Extensive experimental results illustrate that our approach outperforms the state-of-the-art methods. Moreover, we also demonstrate the practical value of our approach to the task of sketch-based image retrieval.

\section*{Acknowledgements}
The work was supported in part by the National Science Foundation of
China (61772158, 61702136, and 61701273), China Postdoctoral Science Foundation (2020M681961), Research Program of Zhejiang Lab (2019KD0AC02 and 2020KD0AA02) and Australian Research Council (ARC) grant (DP150104645).

\bibliography{refs}

\end{document}